\documentclass{article}

\usepackage{arxiv}

\usepackage[utf8]{inputenc} 
\usepackage[T1]{fontenc}    
\usepackage{hyperref}       
\usepackage{url}            
\usepackage{booktabs}       
\usepackage{amsfonts}       
\usepackage{nicefrac}       
\usepackage{microtype}      
\usepackage{graphicx}
\graphicspath{ {./images/} }
\usepackage{amsmath}
\usepackage{caption}

\usepackage{algorithm}
\usepackage{algpseudocode}

\title{Giving Simulated Cells a Voice: Evolving Prompt-to-Intervention Models for Cellular Control.}

\author{
  Nam H. Le \\
  University of Vermont \\
  Vermont, USA \\
  \texttt{namlehai90@gmail.com}
  \And
  Patrick Erikson \\
  Tufts University \\
  Boston, USA
  \And
  Yanbo Zhang \\
  Tufts University \\
  Boston, USA
  \And
  Michael Levin \\
  Tufts University \\
  Boston, USA
  \And
  Josh Bongard \\
  University of Vermont \\
  Vermont, USA
}

\begin{document}
\maketitle
\begin{abstract}
Guiding biological systems toward desired states, such as morphogenetic outcomes, remains a fundamental challenge with far-reaching implications for medicine and synthetic biology. While large language models (LLMs) have enabled natural language as an interface for interpretable control in AI systems, their use as mediators for steering biological or cellular dynamics remains largely unexplored. 
In this work, we present a functional pipeline that translates natural language prompts into spatial vector fields capable of directing simulated cellular collectives. Our approach combines a large language model with an evolvable neural controller (Prompt-to-Intervention, or P2I), optimized via evolutionary strategies to generate behaviors such as clustering or scattering in a simulated 2D environment. We show that even with a limited vocabulary and simplified cell models, evolved P2I networks can successfully align cellular dynamics with user-defined goals expressed in plain language. This work offers a complete loop from language input to simulated bioelectric-like intervention to behavioral output, providing a foundation for future systems capable of natural language-driven cellular control.

\vspace{1em}
\noindent\textbf{Note:} This is the authors’ preprint version. The final version will appear in the ACM GECCO Workshop Proceedings 2025.
\end{abstract}


\section{Introduction}
Complex systems remain inherently difficult to predict and control \cite{hoel2016can}. Biological systems, despite being notoriously complex and thus hard to control, have an advantage over inorganic complex systems in that their behavior can be modified by a wide range of interventions across physical modalities (electrical, mechanical, chemical, vibrations, thermal, optical) -- applied across multiple scales, from molecular interactions to whole-body and even ecosystem-level influences. However, traditional bottom-up approaches to biological control have predominantly focused on molecular mechanisms at the cellular level, often neglecting the emergent behaviors that arise from the collective interactions of cells. Recent insights from information theory and control systems suggest that higher levels of organization may provide effective control points, where emergent dynamamics serve as interfaces for top-down regulation \cite{hoel2020emergence}. For instance, collective behaviors such as those observed in wound healing, morphogenesis, and metastasis are driven by localized interactions yet give rise to robust global patterns essential for life \cite{pezzulo2016top, levin2013reprogramming}. Bridging these hierarchical levels of control remains a key challenge for understanding and influencing emergent biological dynamics.

One way to discover the best angles from which to apply successful interventions to biological targets is to not presuppose the scale and type of intervention. Rather, it could be efficacious to train AI to autonomously discover the appropriate scale and type of intervention free of human investigator bias, as well as which intervention of that type, and at that scale, yield the desired result. 

It is very likely that different desired results will require different interventions of different kind and scale. Thus it would be desirable to (1) develop an interface that allows investigators to rapidly articular and specify desied outcome; (2) employ an AI model that could translate these desired outcomes into interventions likely to bring them about; and (3) provide a complementary interface that renders the resulting system behavior in human-interpretable form. 

In this paper, we present a framework that embodies these ideas in a simulated environment. Our approach—termed \textit{ZapGPT}—leverages natural language as the medium for interfacing with cellular dynamics. Specifically, natural language prompts (e.g., “Cluster!”) are translated into spatially distributed interventions via a Prompt-to-Intervention (P2I) model, and the ensuing cellular dynamics are interpreted back into natural language using a Dynamics-to-Response (D2R) model based on a video language model. While our experiments focus on a simplified task with a constrained set of interventions and outcomes, they serve as an initial demonstration of a language-guided, evolutionary framework for influencing decentralized behaviors in synthetic systems.

We further explore how evolutionary algorithms can be integrated to optimize the mapping from human prompts to cellular interventions. In our study, we focus on simple collective behaviors—such as clustering and scattering—providing a \textit{proof-of-concept} that lays the groundwork for future investigations into more complex and nuanced language-to-cellular dynamics transformations.

\section{Background and Related Work}
\label{sec: rl}

Recent advances in developmental biology, bioelectricity, and multi-scale physiology suggest that biological systems exhibit collective intelligence, integrating information across multiple scales to make adaptive decisions \cite{levin2021bioelectric, levin2022collective}. Cells communicate through bioelectric, biochemical, and mechanical signals, dynamically shaping morphogenesis, regeneration, and behavior. Understanding and harnessing these properties require a framework that enables high-level communication with cellular collectives rather than micromanaging molecular mechanisms.

It has been shown recently that biological systems process information beyond the molecular scale, suggesting that higher-level interventions can shape collective behavior without requiring direct genetic or molecular manipulations \cite{pezzulo2016top, fields2020morphological}. Bioelectric signaling, for instance, provides a powerful mechanism for influencing cellular decision-making, enabling the reprogramming of anatomical structures and regenerative outcomes \cite{levin2019computational, levin2021bioelectric}. Rather than focusing solely on gene editing or biochemical pathways, researchers have explored methods for modulating bioelectric gradients to guide self-organization and pattern formation in tissues \cite{pezzulo2016top, levin2013reprogramming}. 
This shift from reductionist molecular control to emergent, system-level regulation suggests the need for novel computational frameworks that can translate high-level human intent into biologically meaningful interventions. Rather than prescribing specific molecular changes, such frameworks should enable guided self-organization, allowing tissues to achieve target structures or behaviors based on interpretable external cues. 

Similar principles have been demonstrated in synthetic systems, where decentralized agents evolve complex behaviors through local interactions and selection. For example, evolving neuronal controllers for Boid-like agents has shown that coordinated group motion and adaptive dynamics can emerge from simple, bottom-up rules, without centralized control or explicit objectives \cite{le2024emergent}. This line of work underscores the potential of evolutionary computation to generate lifelike, adaptive behaviors — offering a synthetic parallel to the self-organizing principles observed in biological morphogenesis.

Advances in artificial intelligence (AI) and machine learning (ML) have enabled new ways to analyze and control complex systems, including biological networks. Deep learning models have been applied to tasks such as protein folding \cite{xu2019distance}, cellular image analysis \cite{moen2019deep}, and bioelectric signal interpretation \cite{kalaivani2023prediction}. However, most AI applications in biology remain focused on data-driven pattern recognition rather than active control. 

Beyond predictive modeling, evolutionary algorithms (EAs) offer a robust framework for optimizing biologically meaningful interventions, experimental protocols, and control strategies for cellular behaviors \cite{stanley2019designing, bongard2006resilient}. EAs have been applied in fields such as synthetic biology \cite{ray1993evolutionary, cao2010evolving} and bioelectric pattern optimization \cite{hazan2022exploring}, providing a mechanism for tuning system-level parameters to achieve desired outcomes without requiring explicit mechanistic models.

Yet, a key challenge remains: while evolutionary algorithms can optimize interventions, they do not inherently provide an intuitive way for human users to specify goals or interpret results. Most AI-driven approaches in biological control either rely on rigid mechanistic models or produce opaque, uninterpretable policies. To address this, our work proposes leveraging natural language as an intuitive interface for specifying objectives and evaluating outcomes.

Language-based interventions allow for adaptive, dynamic communication between human intent and complex systems, bridging the gap between abstract symbolic goals and physical interventions. Inspired by this idea, our proposed framework integrates large language models (LLMs) to generate structured interventions and vision-language models (VLMs) to interpret the resulting system dynamics. Furthermore, evolutionary algorithms are employed to refine the translation process from natural language to intervention. It is important to note that our approach is presented as a preliminary, proof-of-concept demonstration within a constrained simulation setting, with the ultimate goal of scaling to more complex, real-world scenarios.

The following section describes the key components of this framework and the experimental setup used to evaluate its effectiveness.

\section{Experimental Methods}
\label{sec: exp}

In this section we provide an overview of our approach, followed by a description of each component of the ZapGPT pipeline (Fig. \ref{fig:pipeline}). Our experiments are conducted in a controlled simulated environment and serve as a proof-of-concept for using language-guided, evolutionary optimization to influence system behavior.

\subsection{Framework Overview}
\label{methOverview}

The proposed framework establishes a bidirectional pipeline for translating high-level human intent into actionable interventions and interpreting system dynamics into human-understandable symbolic descriptions. Although the framework is designed to be generalizable to a wide range of complex systems, in this study we apply it to a simplified setting.

\begin{figure*}[ht]
    \centering
    \includegraphics[width=1.0\columnwidth, height=0.3\textheight]{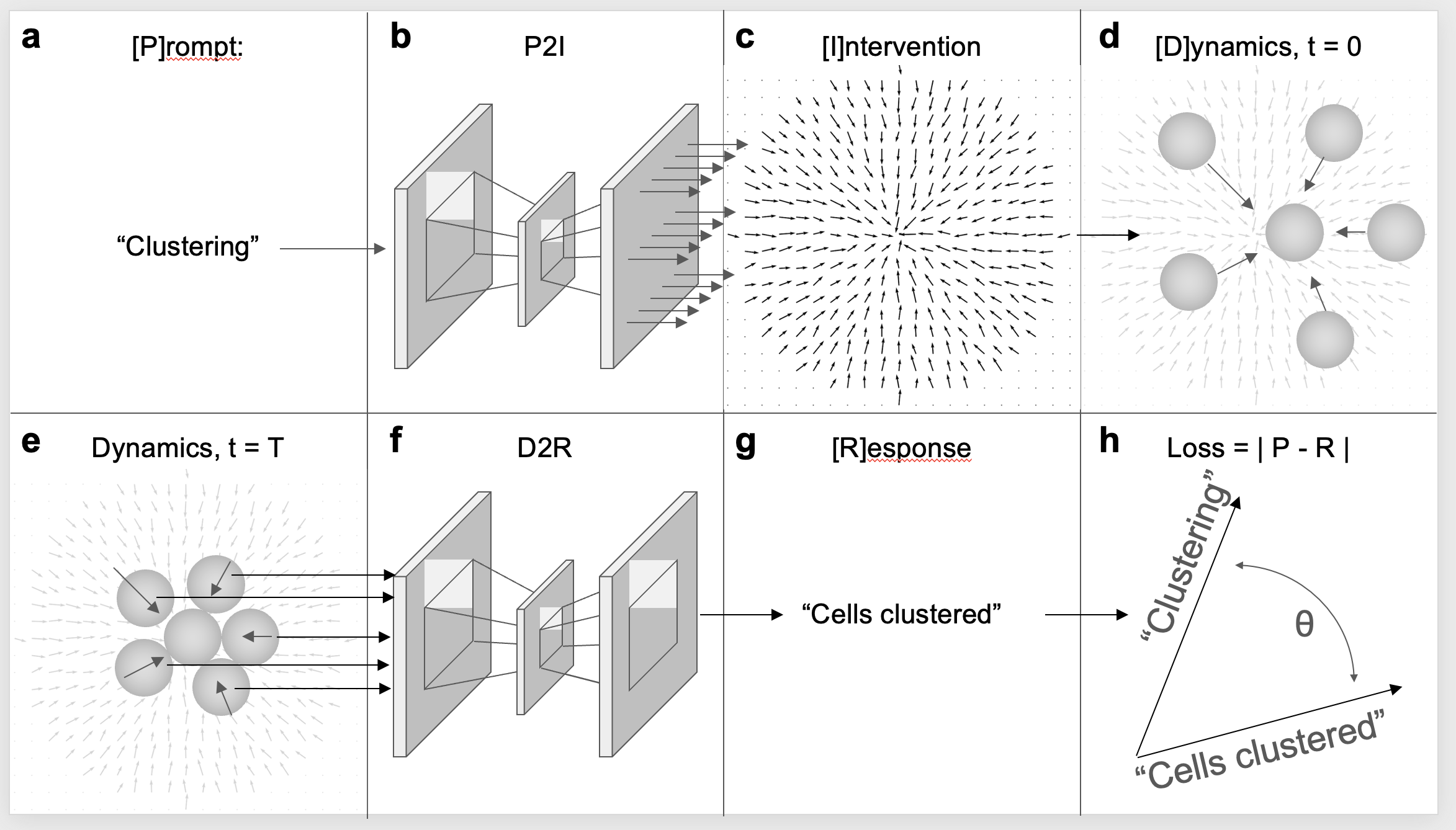}
    \caption{\small Overview of \textbf{ZapGPT}.} 
    \label{fig:pipeline}
\end{figure*}

The pipeline begins with a natural language \textit{prompt}, such as \textit{“achieve clustering”} (Fig. \ref{fig:pipeline}a), representing the desired outcome for the system. This input is input to the \textit{Prompt-to-Intervention (P2I)} model, which uses a language model to generate system-specific interventions (e.g., vector fields for spatially distributed systems (Fig. \ref{fig:pipeline}b)). These interventions influence the complex system (Fig. \ref{fig:pipeline}c) by guiding its initial state and shaping its dynamics over time (Fig. \ref{fig:pipeline}d).

Subsequently, the \textit{Dynamic-to-Response (D2R)} model (Fig. \ref{fig:pipeline}e) processes the system dynamics and produces a human-interpretable symbolic response (e.g., \textit{“system achieved clustering”} (Fig. \ref{fig:pipeline}f). Finally, the pipeline computes a \textit{semantic loss} metric that quantifies the alignment between the original prompt and the generated response (Fig. \ref{fig:pipeline}g), enabling iterative optimization of the P2I model.

This framework highlights two key interfaces: (1) translating human intent into actionable interventionsinterventions through the P2I model, and (2) interpreting system dynamics into symbolic responses through the D2R model. The complex system in between can take various forms, with interventions and behaviors adapted to the domain and the specific problem being addressed. Together, these components provide a scalable method for aligning high-level human intent with the dynamics of diverse systems. In this proof-of-concept study, we restrict our experiments to a simplified set of interventions and responses, thereby focusing on the feasibility of the overall approach.

\subsection{Simulated Environment}

To implement the pipeline, we conducted experiments in a simulated environment designed to study cellular behavior in response to language-guided vector fields. The environment is a continuous 2D rectangular space ($500 \times 500$ units) with reflective boundaries to ensure cells remain within the simulation area.

The environment contains $N=100$ circular cells (radius$r=5$ units), initialized with random positions and velocities. Cells are influenced by external vector fields, represented as $n \times n$ grids (with $n = 2, 3, 5, 10$). Each grid cell contains a vector that defines the direction and magnitude of the force acting on cells in that region. Interpolated field vectors dynamically guide cell movement, overriding initial trajectories.

At each time step, cell positions are updated based on the vector field forces, reflective boundary conditions, and local repulsion to prevent overlap. The simulation records metrics such as the average distance of cells and tracks cell positions over time, which serve as quantitative measures for evaluating the system's response to the interventions.

\subsection{Prompt-to-Intervention (P2I) Model Architecture}

\begin{figure}[ht]
    \centering
    \includegraphics[height=0.5\textheight, keepaspectratio]{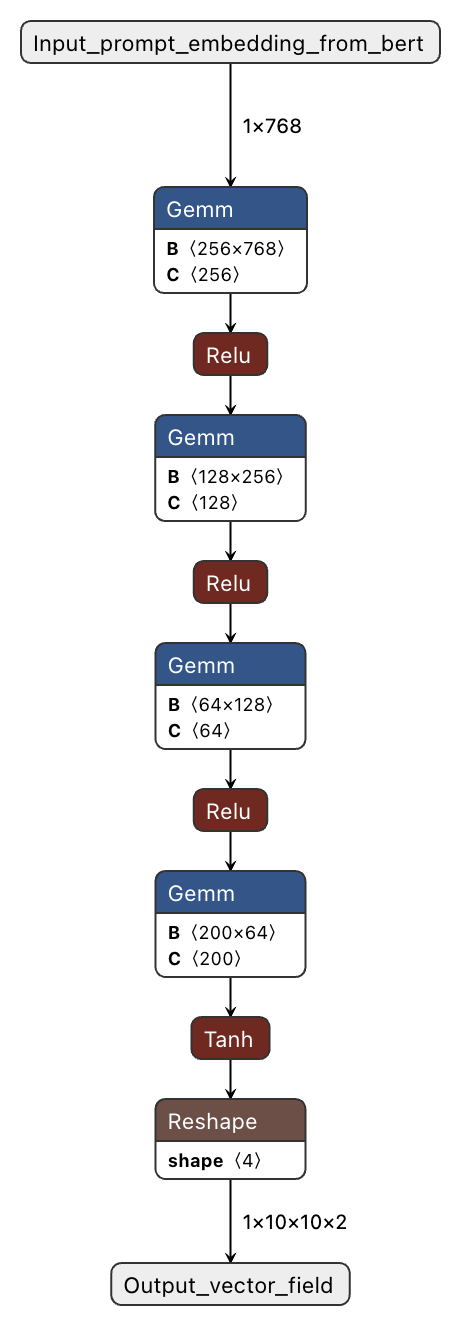}
    \caption{\small The \textit{Prompt-to-Intervention (P2I)} model architecture.} 
    \label{fig:p2i_model}
\end{figure}

The \textit{P2I} model serves as an interface that translates natural language prompts into spatially actionable vector fields -- i.e., interventions that drive cellular dynamics. As illustrated in Figure \ref{fig:p2i_model}, the model begins by encoding the input prompt using a precomputed BERT embedding, yielding a 768-dimensional vector. This embedding is then processed through a feed-forward neural network (FFN) that transforms the input into a flattened  representation of the vector field. The output is reshaped into a 3D tensor with dimensions \((n, n, 2)\), where \(n \times n\) denotes the resolution of the grid, and 2 represents the magnitude and direction of the vector in each grid cell.

By directly generating spatial interventions from natural language, the P2I model establishes a bridge between symbolic instructions and cellular dynamics, forming a foundational component of the ZapGPT framework.

\subsection{Representing System Dynamics for Evaluation}

To evaluate the performance of the P2I-D2R system, we focus on two key aspects of system dynamics: temporal progression and final spatial configuration.

The first criterion, capturing \textit{immediate} or temporal dynamics, is measured by the average pairwise distance metric:
\begin{equation}
D_{\text{avg}}(t) = \frac{1}{N(N-1)} \sum_{i=1}^{N} \sum_{j=i+1}^{N} | \mathbf{x}_i(t) - \mathbf{x}_j(t) |,
\end{equation}
where N is the number of cells, and $\mathbf{x}_i(t)$ denotes the position of the i-th cell at time step t. This metric tracks whether cells move closer together or spread apart over time, providing a quantitative measure of clustering or scattering trends.

The second criterion evaluates the final configuration of cells at the simulation's end, determining whether the cells have formed a cohesive cluster or remain dispersed.

Both metrics are crucial for interpreting system behavior in ways that align with human expectations. Furthermore, advances in vision-language models (VLMs) enable the translation of these quantitative metrics into concise natural language descriptions, facilitating a more intuitive evaluation of the interventions.

\subsection{D2R and Fitness Evaluation}

The \textit{Dynamics-to-Response (D2R)} model evaluates the alignment between cellular behaviors and the original language prompt. To achieve this, it leverages a pretrained vision-language model (VLM), \textit{Moondream2} \cite{vik_2024}, which interpret simulation outputs (e.g., time-series plots and spatial configurations) and generate concise textual responses. These responses, such as \textit{"Clustering"} or \textit{"Scattering"}, are then compared with the input prompt to compute fitness scores.

Among various available vision-language models (VLMs)—such as GPT-4 Vision~\cite{openai2024gpt4technicalreport} and LLava~\cite{liu2024llavanext}—we selected Moondream2~\cite{vik_2024} for its lightweight architecture, fast inference, and suitability for structured visual inputs like simulation plots. By constraining its output to a one-word label (e.g., “Clustering” or “Scattering”), Moondream2 integrates well with our binary evaluation framework and enables efficient iteration during training.\footnote{We validated Moondream2’s performance through a testbed evaluation using artificial plots of average pairwise distance trends and final spatial configurations of clustering and scattering behaviors. These tests confirmed the model’s ability to interpret visual data accurately and produce the expected responses. Code and evaluation results will be released at \url{https://github.com/namlehai90/moondream-score-validation}.}

\subsubsection{Restricting Output to One Word}

Our task is defined by simple, one-word outputs (either \textit{"Clustering"} or \textit{"Scattering"}). To ensure the alignment between the input prompt and the D2R output, we measure their similarity using a loss function, such as cosine similarity. However, differences in dimensionality between embeddings can cause inconsistent results, especially in tasks with fixed, straightforward objectives \cite{tessari2024surpassing}.

To mitigate this, we employ prompt engineering to force the D2R output to a single word--directly matching the dimensionality of the input prompt. This allows us to apply binary or categorical loss functions, providing a straightforward and reliable evaluation framework.

\subsubsection{Input to D2R: Visual and Textual Prompts}

\begin{figure}[ht]
    \centering
    \includegraphics[width=0.7\columnwidth, height=8cm, keepaspectratio]{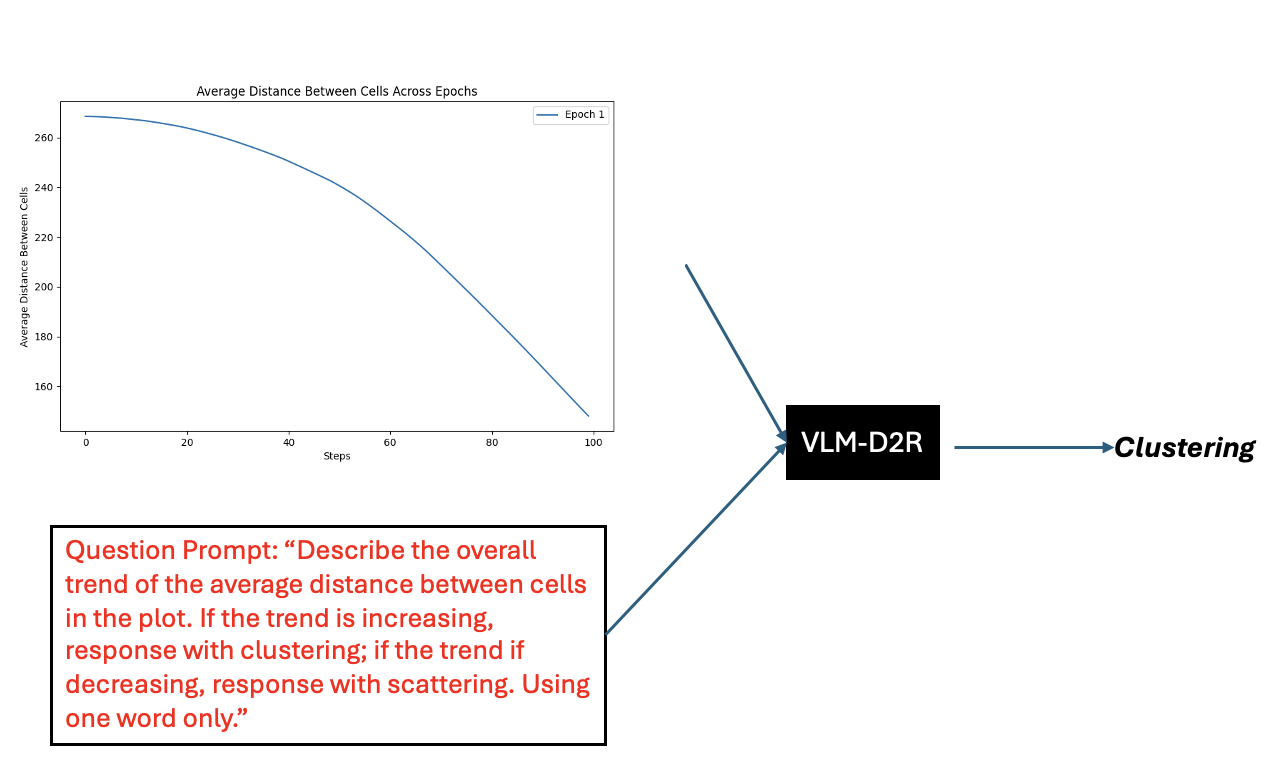}
    \caption{\small D2R input-output example using the average distance plot. The prompt directs the VLM to choose between “clustering” or “scattering” based on the observed trend.}
    \label{fig:distance_D2R}
\end{figure}

\begin{figure}[ht]
    \centering
    \includegraphics[width=0.7\columnwidth, height=8cm, keepaspectratio]{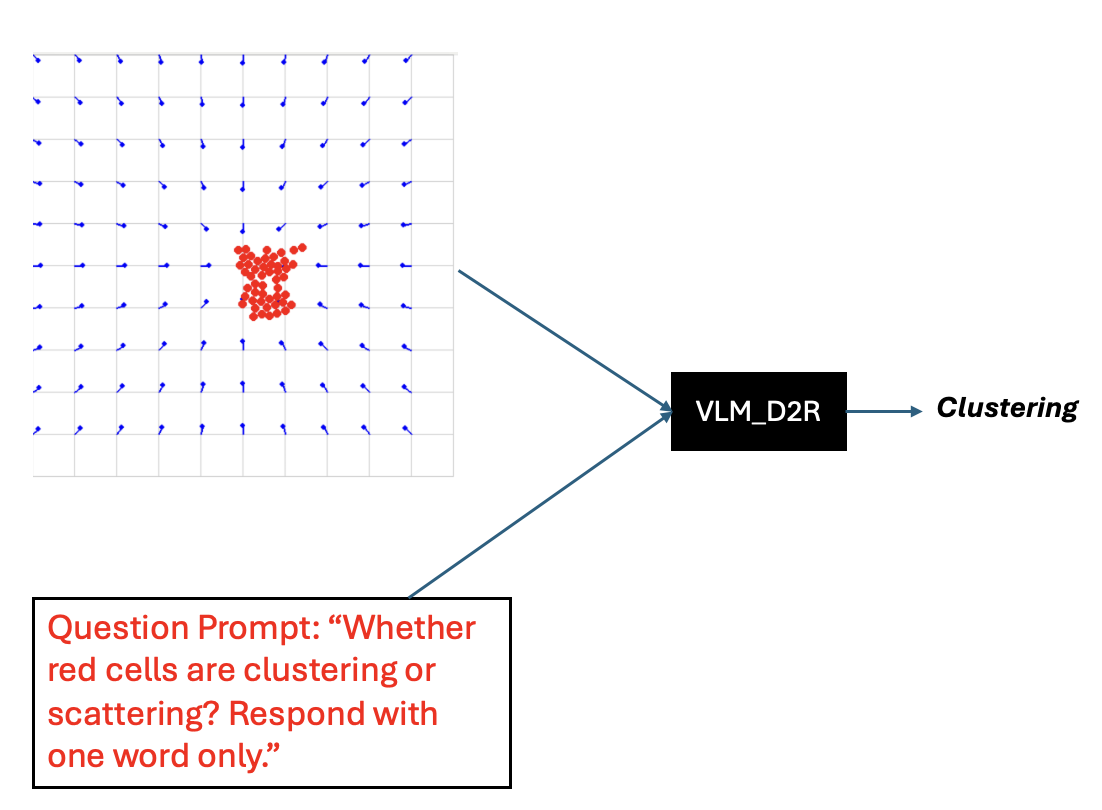}
    \caption{\small D2R input-output example using the final cell position plot. The VLM determines whether the cells exhibit clustering or scattering behavior.}
    \label{fig:position_D2R}
\end{figure}

The D2R model receives two types of inputs: (1) simulation-derived visualizations, and (2) minimal textual prompts to elicit one-word responses. The visual inputs include a time-series distance plot (Figure~\ref{fig:distance_D2R}) and a final cell position plot overlaid with the vector field (Figure~\ref{fig:position_D2R}). Prompts shown in the figures are designed to guide the VLM toward binary classification—“clustering” or “scattering”—which aligns with our evaluation framework.

\subsubsection{Fitness Evaluation}

The fitness evaluation measures the similarity between the original prompt (as processed by the P2I model) and the D2R model's response. In our experimental setup, an \textbf{epoch} is defined as a complete simulation run consistint of the following steps:

\begin{enumerate}
    \item An environment is randomly initialized.
    \item The P2I model translates the input prompt into a corresponding vector field intervention.
    \item The intervention is applied to the environment, and the simulation is run for \textbf{500 steps}, allowing the cellular dynamics to evolve.
    \item At the end of the simulation, the D2R model processes the resulting dynamics and generates a one-word response (e.g., \textit{“Clustering”} or \textit{“Scattering”}).
\end{enumerate}

For each epoch, we assign a binary score \(R_i = 1\) where:

\begin{itemize}
    \item \(R_i = 1\) if the D2R response matches the input prompt, and 
    \item \(R_i = 0\) otherwise.
\end{itemize}

To mitigate randomness and ensure reliable evaluation, we repeat this process over E = 30 epochs and compute the average fitness score as follows:
\begin{equation}
R_{\text{average}} = \frac{\sum_{i=1}^{E} R_i}{E}.
\end{equation}

This approach provides a robust measure of the alignment between the human-specified prompt and the system’s emergent behavior.

For experiments involving both the distance plot and the final position plot, we combine the rewards to balance their contributions:
\begin{equation}
R_{\text{combined}} = \alpha R_{\text{distance}} + \beta R_{\text{position}},
\end{equation}
where \(\alpha = 0.5\) and \(\beta = 0.5\). This combined reward function captures both the temporal dynamics and the final spatial configuration, providing a robust metric for evaluating the alignment between human intent and system behavior.

With this fitness/loss function in place, the next step is to optimize the vector fields generated by the \textit{Prompt-to-Intervention (P2I)} model to better achieve the desired outcomes.

\subsection{Evolutionary Algorithms}

We model our problem as optimizing the P2I model so that it can generate vector fields which, when applied to guide cellular dynamics, yield behaviors that closely match the original input prompt as the intended outcome. In other words, our objective is to find an optimal configuration of the P2I model’s neural network weights so that the resulting cellular dynamics align with the goal (e.g., \textit{'clustering'}). It is important to note that we are not directly optimizing the vector fields themselves; instead, we optimize the P2I model to produce the desired interventions.

This optimization problem is well-suited for evolutionary algorithms (EAs), which are effective in exploring high-dimensional, nonlinear search spaces. Unlike reinforcement learning (RL) methods—often associated with large language models (LLMs)—traditional backpropagation is challenging in our pipeline due to sparse rewards and non-differentiable components (e.g., the emergent dynamics of cellular behavior and the evaluations from the Dynamics-to-Response (D2R) model) \cite{salimans2017evolution, stanley2019designing}. EAs, which rely solely on reward evaluations, thus provide an attractive alternative by bypassing the need for gradient propagation.

\subsubsection{(1+1) Evolution Strategy}

Our initial approach employs the simple \textit{(1+1) Evolution Strategy (ES)} \cite{Ostermeier1994} with an adaptive mutation step size to optimize the P2I model’s weights. This approach is chosen for its intuitive hill-climbing behavior, making it suitable for initial experiments with intermediate rewards (e.g., those based solely on the distance plot).

The \textit{(1+1)-ES} uses mutation as its sole evolutionary operator. Key parameters and the optimization process are detailed in Algorithm~\ref{alg:one_plus_one_es}.

\begin{algorithm}
\footnotesize
\caption{(1+1) Evolution Strategy with Adaptive Step Size}
\label{alg:one_plus_one_es}
\begin{algorithmic}[1]
\Require Initial P2I weights \(W\), mutation step size \(\sigma = 0.1\), target success rate \(p_{\text{target}} = 0.2\), sliding window size \(s = 5\), maximum generations \(G\).
\Ensure Optimized P2I weights \(W^*\).

\State Randomly initialize \(W_{\text{parent}}\).
\State Compute initial fitness \(R_{\text{parent}} = \text{Fitness}(W_{\text{parent}})\).
\State Initialize success window \(\mathcal{S} = \{\}\).

\For{\(g = 1\) to \(G\)}
    \State \textbf{Mutation:} \(W_{\text{offspring}} = W_{\text{parent}} + \mathcal{N}(0, \sigma)\).
    \State \textbf{Evaluate:} \(R_{\text{offspring}} = \text{Fitness}(W_{\text{offspring}})\).
    \If{\(R_{\text{offspring}} \geq R_{\text{parent}}\)}
        \State \(W_{\text{parent}} \gets W_{\text{offspring}}\).
        \State \(R_{\text{parent}} \gets R_{\text{offspring}}\).
        \State Append \(1\) to \(\mathcal{S}\) \Comment{Success}.
    \Else
        \State Append \(0\) to \(\mathcal{S}\) \Comment{Failure}.
    \EndIf
    \If{\(|\mathcal{S}| > s\)}
        \State Remove oldest entry from \(\mathcal{S}\).
    \EndIf
    \State Compute success rate \(p_{\text{success}} = \frac{\sum \mathcal{S}}{|\mathcal{S}|}\).
    \If{\(p_{\text{success}} > p_{\text{target}}\)}
        \State \(\sigma \gets \sigma \times 0.9\) \Comment{Reduce step size}.
    \Else
        \State \(\sigma \gets \sigma \times 1.1\) \Comment{Increase step size}.
    \EndIf
    \State Clamp \(\sigma\) to \([1\text{e-6}, 5.0]\).
\EndFor
\State \Return \(W_{\text{parent}}\) as \(W^*\).
\end{algorithmic}
\end{algorithm}

While (1+1)-ES demonstrated good performance with simpler vector field configurations (e.g., $2 \times 2$ and $3 \times 3$), our experiments showed that it struggled to achieve optimal clustering behavior in more complex settings (e.g., $5 \times 5$ and $10 \times 10$ grids). This limitation motivates the use of more robust optimization methods.

\subsubsection{Genetic Algorithms}

For larger search spaces, we implemented a real-valued genetic algorithm (GA) with arithmetic crossover \cite{holland1992genetic} to optimize the P2I model’s weights. The key components of our GA are as follows:

\begin{itemize}
\small
\item \textbf{Pop\_size = 20:} Twenty P2I weight candidates are randomly generated as the initial population.
\item \textbf{Selection:} Tournament selection is performed with a tournament size of 8.
\item \textbf{Crossover:} Arithmetic crossover is applied to generate offspring:
    \[
    W_{\text{child}} = \alpha W_{\text{parent1}} + (1 - \alpha) W_{\text{parent2}},
    \]
    where \(\alpha \in [0, 1]\) is chosen randomly.
    \item \textbf{Mutation:} Gaussian noise is added to weights:
    \[
    W_{\text{mutated}} = W_{\text{original}} + \mathcal{N}(0, \sigma),
    \]
    with \(\sigma = 0.1\).
    \item \textbf{Fitness Evaluation:} Each individual is evaluated using the combined reward function derived from both the distance plot and the final position plot.
\end{itemize}

The GA iteratively performs selection, crossover, mutation, and fitness evaluation for a 50 generations. By maintaining a diverse population and utilizing crossover operations, the GA effectively explores larger search spaces and overcomes some of the limitations observed with (1+1)-ES.

\section{Experimental Results}

\subsection{Evaluating (1+1)-ES with Single Criterion (Immediate Reward)}

We begin by assessing the simplest setting, where we use \textit{(1+1)-ES} to optimize the P2I model with only an immediate reward--derived from the average distance plot. Each vector field configuration ($2 \times 2$, $3 \times 3$, $5 \times 5$, $10 \times 10$) was evaluated over 30 random seeds, and the results were averaged to mitigate randomness. \footnote{For a visual demonstration of the ES and GA optimization processes in ZapGPT, please refer to the accompanying video at \href{https://www.youtube.com/watch?v=t2juNHTRs-o&list=PLchLYCCaLyQcM4OGr9Nsxe-ycN-QzWW7A&index=1}{this link}.}

To monitor optimization progress, we report the best fitness values over generations for each configuration. Additionally, we conducted a wilcoxon signed-rank test comparing the fitness at  generation 0 with that at the last generation to assess the statistical significance of the improvements.

\begin{figure*}[ht]
    \centering
    \includegraphics[width=\textwidth, height=0.7\textheight]{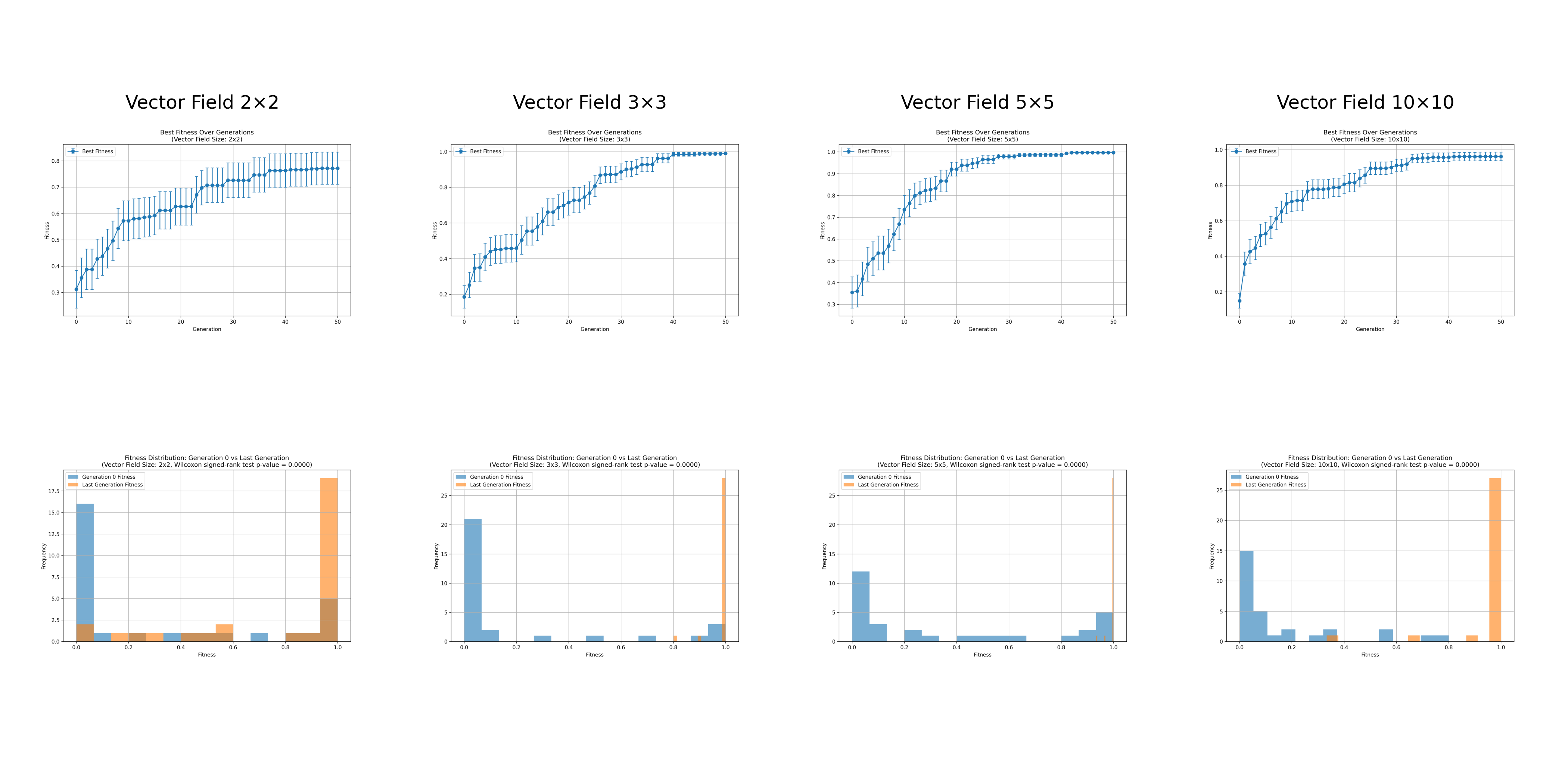}
    \caption{\small Best fitness scores across generations for different vector field sizes, using (1+1)-ES and average distance metric.}
    \label{fig:fitness_progression}
\end{figure*}

Across all vector field sizes, (1+1)-ES consistently improves fitness over generations, as shown in Figure~\ref{fig:fitness_progression}. The Wilcoxon signed-rank test (p-value < 0.05) confirms that the final fitness distributions are statistically significantly better than the initial ones. These results indicate that, when evaluated solely on the distance plot, (1+1)-ES can effectively optimize the clustering behavior in our simulated environment.

However, while these results are promising for the immediate reward criterion, they do not necessarily guarantee that the cellular dynamics will converge to a single cohesive cluster. This observation motivates the evaluation of an additional criterion.

\subsection{Evaluating (1+1)-ES with Combined Rewards}

To better capture both the progression of cellular behavior and the final spatial arrangement, we extended the evaluation to include a combined reward. This combined reward function comprises: 

\begin{itemize}
\item An \textit{immediate reward} based on the distance plot, which encourages cells to move closer together over time.
\item A \textit{final reward} based on the final spatial configuration, ensuring that cells form a single cohesive cluster rather than multiple small groups.
\end{itemize}

\begin{figure*}[ht]
\centering
\includegraphics[width=\textwidth, height=0.7\textheight]{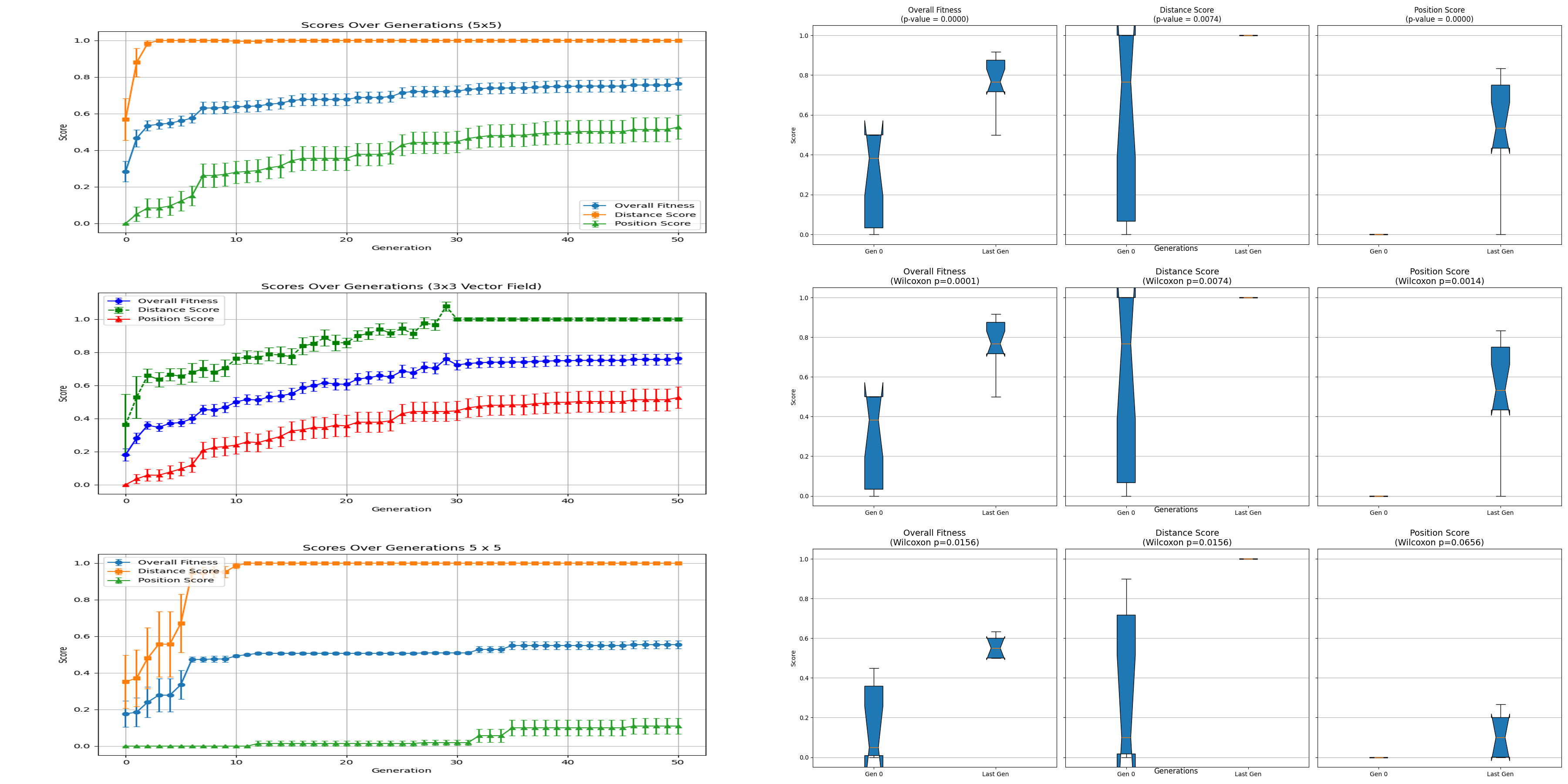}
\caption{\small Scores over generations for different vector field sizes using \textit{(1+1)-ES} with combined rewards.}
\label{fig:combined_fitness_progression}
\end{figure*}

We tested vector field settings of $2 \times 2$, $3 \times 3$, and $5 \times 5$ over 30 random seeds. Figure~\ref{fig:combined_fitness_progression} illustrates the evolution of the combined fitness scores, including error bars that capture the variability across runs. Statistical testing again confirmed that the improvements from generation 0 to the final generation were significant (p < 0.05).

The results reveal that for smaller grids ($2 \times 2$ and $3 \times 3$), (1+1)-ES successfully balances both criteria, leading to the formation of a single cohesive cluster. However, for the $5 \times 5$ grid, although the overall fitness improves, the final spatial configuration (position score) remains suboptimal. This suggests that while (1+1)-ES can reduce pairwise distances effectively, it struggles to enforce global cohesion in more complex scenarios.


\subsection{Genetic Algorithms on 5x5 and 10x10 vector fields}

Building on the limitations observed with (1+1)-ES in higher-dimensional settings, we investigated a Genetic Algorithm (GA) as an alternative. In our GA experiments, we focus on configurations where (1+1)-ES underperforms, specifically for $5 \times 5$ and $10 \times 10$ vector fields.

\begin{figure*}[ht]
    \centering
    \includegraphics[width=\textwidth]{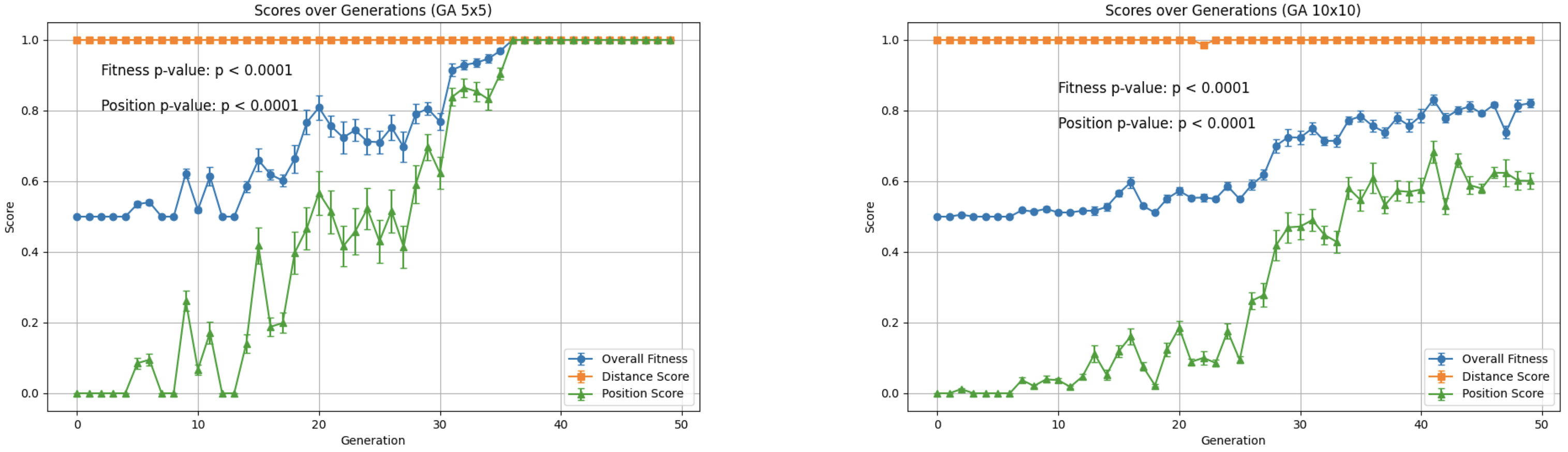}
    \caption{\small Fitness over generations using GAs for different vector field sizes}
    \label{fig:ga_fitness_combined}
\end{figure*}

Using the same combined reward function, we tracked overall fitness, distance scores, and position scores over generations. As shown in Figure~\ref{fig:ga_fitness_combined}, GA-driven evolution improved both the overall fitness and the position scores more effectively than (1+1)-ES. Notably, in the $10 \times 10$ setting, the position score steadily increased, suggesting that the crossover mechanism in the GA helps the population explore solutions that balance both local proximity and global cohesion.

Statistical tests comparing the initial and final generations yielded p-values < 0.0001, indicating a significant improvement in both metrics. Although these results are preliminary, they highlight the potential of GA-based optimization to overcome the limitations of local search methods like (1+1)-ES in more complex vector field configurations.

A key advantage of GAs is evident in the larger vector field setting ($10 \times 10$), where the position score increases steadily, indicating an ability to achieve a well-structured cluster compared to $(1+1)$-ES. This suggests that crossover mechanisms here contribute to a more effective search process, helping the population discover solutions that balance both local and global structure.

\textbf{Summary:} Our experimental results demonstrate that while (1+1)-ES is effective in simple settings and for optimizing immediate rewards, its performance deteriorates when the task requires balancing multiple objectives in higher-dimensional spaces. In contrast, Genetic Algorithms show a clear advantage in these more challenging scenarios by effectively guiding both local and global aspects of cellular dynamics. These findings establish a preliminary proof-of-concept for our language-guided, evolutionary framework, while also outlining the need for further research to address scalability and robustness in more complex systems.

\section{Discussion and Future Directions}

\subsection{Applying P2I-D2R to Real-World Systems}

While our current work is based on a simplified simulation, the principles of the P2I-D2R framework have broader applicability:

\paragraph{Real-Cell Biology Experiments:}
In biological research, understanding cellular responses to environmental interventions is crucial. A P2I-driven system could generate optimized intervention schedules (e.g., for releasing chemical signals or drugs) in a bioreactor. The resulting cellular dynamics, such as changes in intracellular signaling, can be recorded and analyzed by D2R to inform further optimization \cite{subramanian2013ultrasonic, karoui2022chemical}.

\paragraph{Swarm Robotics:}
In swarm robotics, controlling distributed agents to achieve coordinated behaviors like clustering, dispersing, or coordinated navigation is a significant challenge. The same framework could generate control signals for robotic swarms, with D2R evaluating the emergent formations. This language-guided approach could enable high-level control over complex, dynamic multi-agent systems \cite{schranz2020swarm, trianni2003evolving}.

By demonstrating that language-guided intervention and evaluation can generalize across synthetic and real-world domains, our study highlights the broader implications of integrating natural language interfaces with dynamic system control.

\subsection{Generalization and Task Expansion}

Our experiments demonstrate the feasibility of the ZapGPT framework for steering collective cellular behavior based on natural language prompts. While the present study focuses on a constrained scenario using simple instructions (e.g., \textit{clustering}), it lays the groundwork for translating language to spatial interventions through learned vector fields.

In ongoing extensions, we introduce multi-word prompts (e.g., \textit{clustering slowly}, \textit{scattering quickly}) to explore how the model handles behaviorally distinct but linguistically similar instructions. Although such prompts may be close in embedding space, they often correspond to qualitatively different control strategies in simulation. For instance, “scattering slowly” and “scattering quickly” differ subtly in language but require distinct intervention dynamics—highlighting a disconnect between semantic and behavioral similarity.

\begin{figure}[ht]
\centering
\includegraphics[width=0.7\columnwidth]{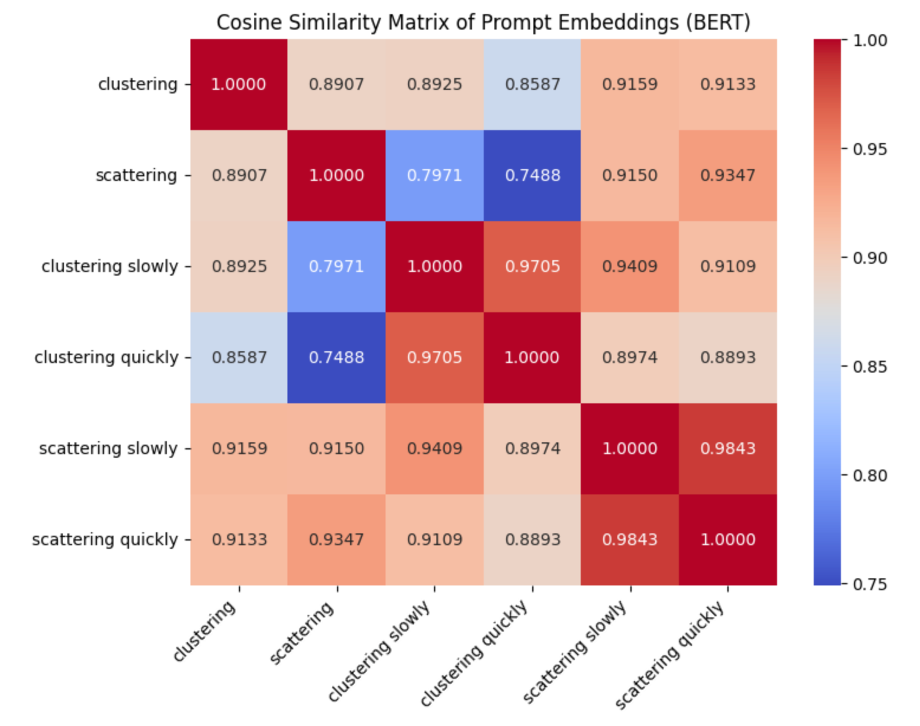}
\caption{\small Cosine similarity matrix of BERT embeddings for different prompts. Linguistic similarity does not imply behavioral similarity.}
\label{fig:prompt_embedding_matrix}
\end{figure}

Training a single P2I model on multiple behaviors—especially those with conflicting goals—can lead to interference, where learning one behavior degrades another. To address this, future work will explore training strategies such as modular architectures, prompt-conditioned layers, and curriculum-based evolution to support multi-behavior learning.

Currently, we use domain-specific metrics (e.g., average intercellular distance) combined with restricted binary captions from a vision-language model (VLM) to evaluate behavioral alignment. This strategy enables clean optimization but depends on constrained language and engineered reward functions, which limit scalability.

To move beyond this, we plan to use flexible, language-grounded evaluators that allow both prompts and outputs to take natural language form. One approach involves using a VLM such as Moondream to generate a free-form caption of the simulation outcome, followed by a second model (e.g., Ollama or DeepSeek-R1) that compares the caption to the prompt. Another strategy leverages CLIP \cite{radford2021learning}, framing the prompt as a text input and the simulation snapshot as an image, using CLIP’s similarity scores to assess alignment.

These alternatives eliminate the need for handcrafted priors but introduce a new challenge: ensuring that semantic similarity reflects behavioral success. For example, if a prompt says “form a cluster” but the VLM caption is “cells are spreading out,” some embedding-based methods may still assign a high similarity score, leading to misleading optimization signals. Designing evaluators that are both behaviorally accurate and semantically meaningful is a key direction for future work.

With improved scoring, we aim to test generalization across semantically diverse instructions. For instance, the model could be trained on “form a cluster” and tested on variants such as “group together” or opposites like “spread apart.” This allows us to examine not only success but also whether the model internalizes the structure of natural language and maps it to behavior accordingly.

Lastly, while this paper centers on short, abstract prompts under the assumption that they are easier to learn, they may in fact present greater ambiguity due to their lack of context. Richer prompts (e.g., “form a tight group near the center”) provide more linguistic structure and may lead to faster and more robust learning. Future work will explore reversing the learning order—training on descriptive prompts first, then testing generalization to abstract forms like “cluster.” This mirrors how humans often ground abstract language through experience before understanding symbolic commands.

Together, these directions will advance ZapGPT toward more expressive, generalizable, and interpretable control of complex collective systems using natural language.

\bibliographystyle{plain}
\bibliography{zapgpt-v1}

\end{document}